\documentclass[conference]{IEEEtran}
\IEEEoverridecommandlockouts
\usepackage{cite}
\usepackage{amsmath,amssymb,amsfonts}
\usepackage{algorithmic}
\usepackage{multirow}% new
\usepackage{algorithm}% new
\usepackage{graphicx}
\usepackage{textcomp}
\usepackage{xcolor}
\def\BibTeX{{\rm B\kern-.05em{\sc i\kern-.025em b}\kern-.08em
    T\kern-.1667em\lower.7ex\hbox{E}\kern-.125emX}}
\begin{document}

\title{Improving Multi-Domain Task-Oriented Dialogue System with Offline Reinforcement Learning}

\author{\IEEEauthorblockN{1\textsuperscript{st} Dharmendra Prajapat}
\IEEEauthorblockA{\textit{Dept of Computer Science and Engineering} \\
\textit{Indian Institute of Technology}\\
Roorkee, India \\
dharmendra\_p@cs.iitr.ac.in}
\and
\IEEEauthorblockN{2\textsuperscript{th} Durga Toshniwal}
\IEEEauthorblockA{\textit{Dept of Computer Science and Engineering} \\
\textit{Indian Institute of Technology}\\
Roorkee, India \\
durga.toshniwal@cs.iitr.ac.in}
}

\maketitle

\begin{abstract}
Task-oriented dialogue (TOD) system is designed to accomplish user-defined tasks through dialogues. The TOD system has progressed towards end-to-end modeling by leveraging pre-trained large language models. Fine-tuning the pre-trained language models using only supervised learning leads to the exposure bias and token loss problem and it deviates the models from completing the user's task. To address these issues, we propose a TOD system that leverages a unified pre-trained language model, GPT2, as a base model. It is optimized using supervised learning and reinforcement learning (RL). The issues in the TOD system are mitigated using a non-differentiable reward function. The reward is calculated using the weighted sum of the success rate and BLEU evaluation metrics. The success rate and BLEU metrics in reward calculation guide the language model for user task completion while ensuring a coherent and fluent response. Our model is acquired by fine-tuning a pre-trained model on the dialogue-session level which comprises user utterance, belief state, system act, and system response. Experimental results on MultiWOZ2.1 demonstrate that our model increases the inform rate by 1.60\% and the success rate by 3.17\% compared to the baseline.
\end{abstract}

\begin{IEEEkeywords}
Pre-trained language model, Natural response generation, 
Conversational AI, GPT-2  
\end{IEEEkeywords}

\section{Introduction}
A task-oriented dialogue (TOD) system aims to achieve a particular task by involving in a dialogue with a user. The TOD systems have applications in multiple fields such as customer services, plane ticket booking, hotel booking, etc~\cite{seneff2000dialogue}. The TOD systems can function as a virtual assistant by completing specific tasks without human intervention~\cite{ram2018conversational}. 

Traditionally, a task-oriented dialogue system is acquired by a pipeline approach~\cite{young2013pomdp} which involves four separate modules.
\begin{figure}[h]
    \centering
    \includegraphics[height =7.5cm, width=0.485\textwidth]{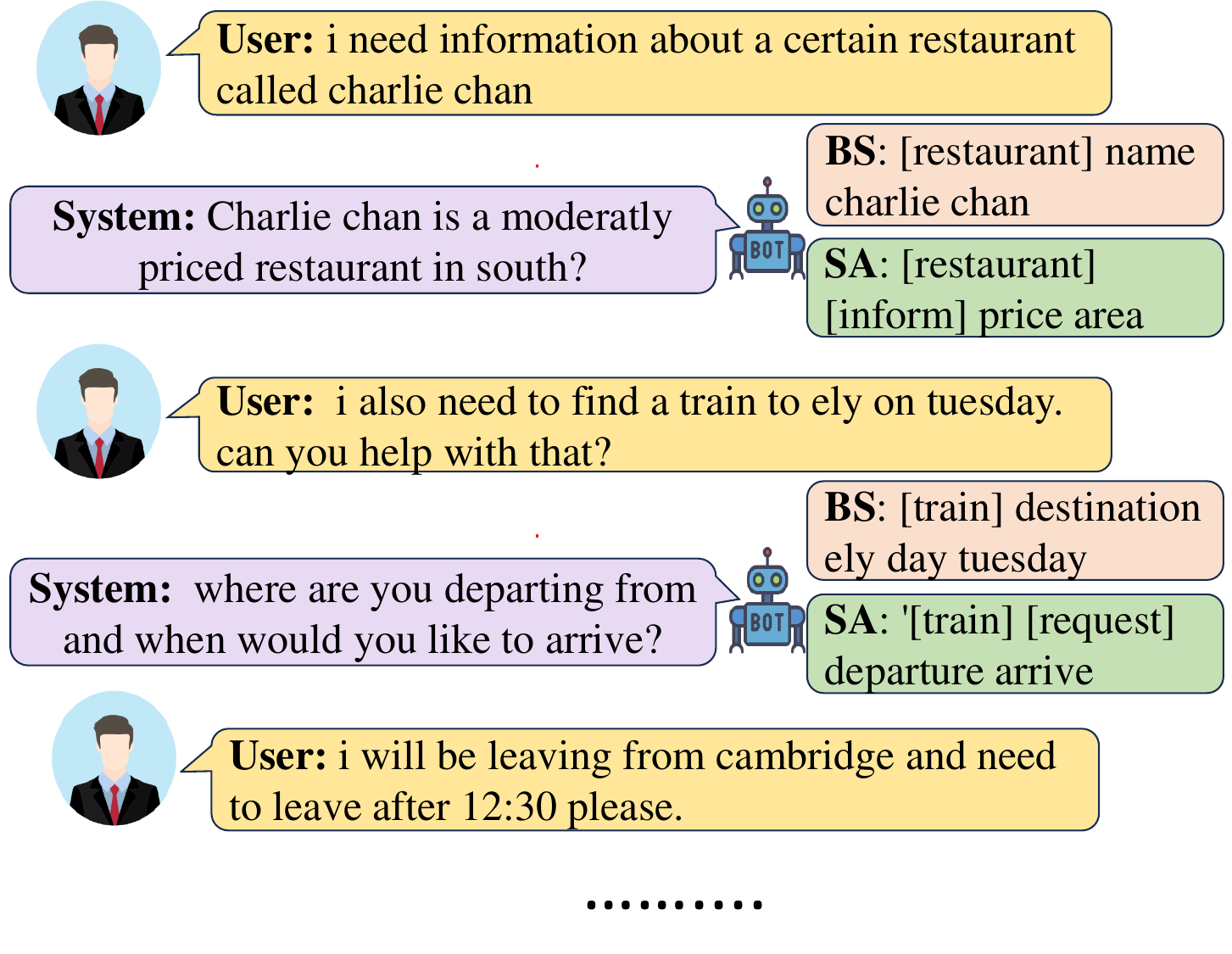} 
    \caption{A multi-domain dialogue between a user and a task-oriented dialogue system}
    \label{fig:intro}
\end{figure}
The natural language understanding (NLU)~\cite{kim2017onenet} module is used to identify the user intent and pre-defined slot (Ex. internet, parking, price range, etc) values. The dialogue state tracking (DST)~\cite{williams2014dialog} module tracks the belief state of users which is represented by pre-defined slots and keeps them updated over each dialogue turn. The system action (SA) module decides the appropriate action to be taken by the system and the natural language generation (NLG)~\cite{wen-etal-2015-semantically} module converts the system actions into coherent and fluent system responses. All these modules are trained and evaluated in a pipeline approach using separate objectives and evolution metrics. This process is cumbersome to train and evaluate the modules separately and it also propagates the error from one module to the subsequent module. A multi-domain dialogue between the user and a TOD system is shown in Figure~\ref{fig:intro} with the generated belief state (BS) and system action (SA).

Recently, task-oriented dialogue (TOD) systems have evolved from a pipeline approach to an end-to-end modeling approach. The end-to-end modeling of the TOD system is achieved using sequence-to-sequence (seq-to-seq) models that are either trained from scratch or leverage pre-trained models to accomplish the user tasks. Methods like Sequicity~\cite{lei2018sequicity} and MOSS~\cite{liang2020moss} use the seq-to-seq model to generate belief state and system response. DAMD~\cite{zhang2020task} is trained in an end-to-end setting for generating multiple system responses under the same context. Task-oriented dialogue (TOD) systems have seen significant advancements with the integration of pre-trained large language models (LLMs)~\cite{devlin-etal-2019-bert,raffel2020exploring}. LLMs are fine-tuned on a particular dataset to generate dialogues of the downstream task. LLM-based TOD systems are commonly developed using a unified model instead of multiple modules~\cite{hosseini2020simple,peng-etal-2021-soloist,yang2021ubar,ham2020end}. The unified model is easy to train using a single objective as compared to the pipeline approach which involves separate objectives for each module. Additionally, unified models are easy to evaluate. SimpleTOD~\cite{hosseini2020simple} uses the dialogue turn-level approach and UBAR~\cite{yang2021ubar} uses the dialogue session-level approach to fine-tune the GPT-2 pre-trained model to complete the user's task. To generalize the end-to-end setting of TOD systems, models such as Soloist~\cite{peng-etal-2021-soloist}, SimpleTOD, and UBAR use generated belief states, rather than ground truth belief states, to query the database. These approaches leverage the MultiWOZ dataset, a multi-domain task-oriented dialogue dataset, to fine-tune the pre-trained model on standard modeling tasks. 

Some approaches fine-tune pre-trained models on auxiliary tasks to make better use of the annotated dataset. These auxiliary tasks also help the model to learn general patterns to enhance its performance on the primary text generation task. MTTOD~\cite{lee2021improving} uses span prediction and RSTOD~\cite{cholakov-kolev-2022-efficient} uses response selection as an auxiliary task along with language modeling objective as a primary task and both methods use T5 pre-trained language model as a base model. 

Primarily, end-to-end task-oriented dialogue systems that leverage pre-trained language models rely on supervised fine-tuning for text generation. Such methods for text generation face exposure bias problems and token loss problems. In the exposure bias problem, the model generates text that resembles the training data since it is exposed to ground truth values during the training but relies on self-generated tokens during inference. In the token loss problem, the training objective single-handedly depends on individual word prediction ignoring the generated token sequence. Simply predicting a sequence of tokens one by one does not lead to the performance improvement of the TOD system. These problems deviate the model from the user's task completion. We need to steer the pre-trained model towards the user's task completion. 

To overcome these issues, We use the reinforcement learning approach to improve the task-oriented dialogue system in an end-to-end setting. Specifically, we use the weighted sum of success rate, and BLEU as a reward to direct the pre-trained uni-directional language model GPT-2 to generate the desired token sequence. The reward is based on the success rate of the dialogue so it directs the model to focus on completing the user's task rather than merely predicting the next token. The reinforcement learning approach helps the model to generate more contextually appropriate, task-oriented, and coherent responses, ultimately improving the system's performance in real-world applications. Our approach enables the pre-trained model to adapt to the downstream task by using a reward that is aligned with the user's task completion. 

In summary, our paper has the following contributions:
\begin{itemize}

\item We propose a reinforcement learning way to optimize the unidirectional pre-trained language model for an end-to-end multi-domain task-oriented dialogue system.

\item We propose a reward function that utilizes success rate and BLEU to optimize the model alongside supervised fine-tuning of the pre-trained model.

\item The proposed approach outperforms the baseline method in the end-to-end multi-domain TOD system on the MultiWOZ2.1 dataset. 

\end{itemize}
\section{Related work}
\subsection{Pipeline-based task-oriented dialogue system}
Traditionally, task-oriented dialogue systems are achieved using pipeline-based methods.
This method contains four modules: Natural language understanding (NLU)~\cite{kim2017onenet}, belief state tracker (DST)~\cite{williams2014dialog}, dialogue policy learning (DPL), and natural language generation (NLG)~\cite{wen-etal-2015-semantically} module.
NLU module helps to understand the user intent and also extracts key information useful for belief state tracker. The DST module monitors the constraints set by the user and continuously updates them with each dialogue turn. The DPL module decides the system action based on the belief state and the NLG module converts the system action into fluent and coherent natural language. All modules are trained sequentially which makes it harder to optimize it on a common objective~\cite{liu2018end}. The drawback of this approach is that errors in one module propagate and can amplify in subsequent modules.

\subsection{End-to-end task-oriented dialogue system}
To overcome the limitations of pipeline-based approaches, end-to-end trainable task-oriented dialogue systems have been developed. These systems are optimized with a unified objective making them easier to train. It allows them to map input directly to the output. A seq2seq model is used by Sequicity~\cite{lei2018sequicity} to generate belief spans and responses. DAMD~\cite{zhang2020task} uses separate decoders to generate belief spans and act spans and multiple decoders to generate system responses under the same dialogue context. With the progress in large language models, they are leveraged in an end-to-end trainable TOD system~\cite{hosseini2020simple,ham2020end, peng-etal-2021-soloist, yang2021ubar}. Such models are capable of generating human-level responses because of their exposure to large amounts of data during the pre-training stage. A pre-trained large language model can only be effectively adapted to a TOD system if a substantial amount of data is available for fine-tuning. The pre-trained models have been fine-tuned using different approaches to achieve the TOD system. SimpleTOD~\cite{hosseini2020simple} uses pre-trained GPT2 as a base model and fine-tunes it using dialogue turn-level whereas UBAR~\cite{yang2021ubar} also uses the GPT-2 but fine-tunes it using dialogue session-level. At the dialogue turn level, a user utterance and the corresponding response are used as an input sequence whereas at the dialogue session level, the input sequence includes all intermediate information generated during the session, such as belief states and system acts across all dialogue turns. We adopt the approach used in UBAR to fine-tune the pre-trained language model, GPT-2, for our system.

\begin{figure}[htp]
    \centering 
    \includegraphics[height= 3cm, width=0.483\textwidth]{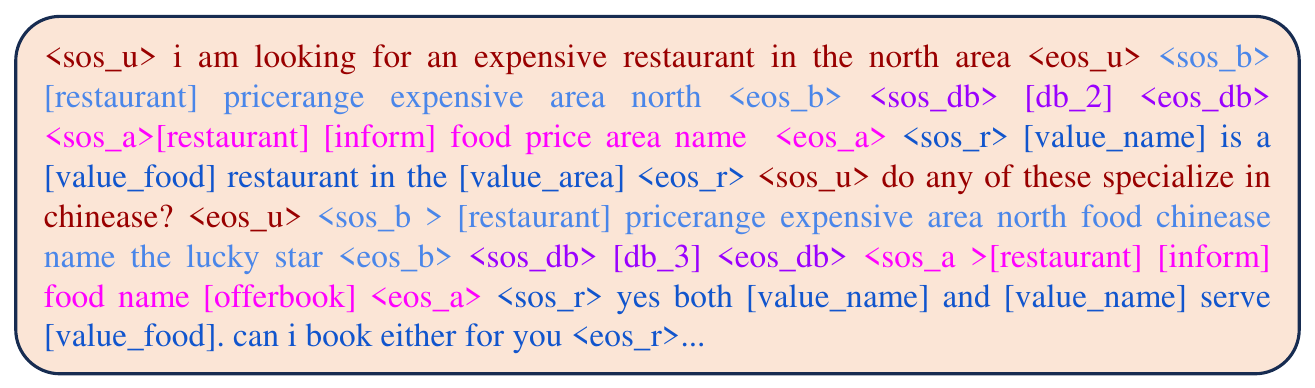} 
    \caption{A dialogue session contains user utterance, belief state, DB search results, system acts, and system response of all dialogue turns and each one is colored differently. }
    \label{fig:dialogue_session}
\end{figure}

Some methods~\cite{he2022galaxy,peng-etal-2021-soloist,liu2021pretraining} train the models from scratch using external dialogue corpora instead of using a pre-trained model however training such models for the TOD system requires the annotated data which limit the data that can be used. GALAXY~\cite{he2022galaxy} uses UniLM, a transformer-based architecture that is pre-trained and fine-tuned on conversational datasets. It uses different types of embeddings such as type and turn embedding along with token and position embedding to improve the representation of the input sequence.

There exist some methods that improve the internal representation of the pre-trained language model by fine-tuning it on additional tasks. MTTOD~\cite{lee2021improving}, and RSTOD~\cite{cholakov-kolev-2022-efficient} use span prediction and response selection as auxiliary tasks respectively. These methods improve the baseline by using large parameterized models such as T5 with 220M parameters. AuGPT~\cite{kulhanek-etal-2021-augpt} uses the data augmentation technique during the pre-training stage and trains the model on multiple auxiliary tasks. Some methods freeze the parameters of the pre-trained model by using the adapters. TOATOD~\cite{bang-etal-2023-task} uses adapters for belief state tracking and natural language generation of the TOD system whereas GPT-ACN~\cite{wang2022task} uses adapters between each of the GPT-2 blocks for end-to-end modeling and uses a coping mechanism at the text generation stage.

\section{Methodology}
\begin{figure*}[h]
    \centering
    \includegraphics[width=1\textwidth]{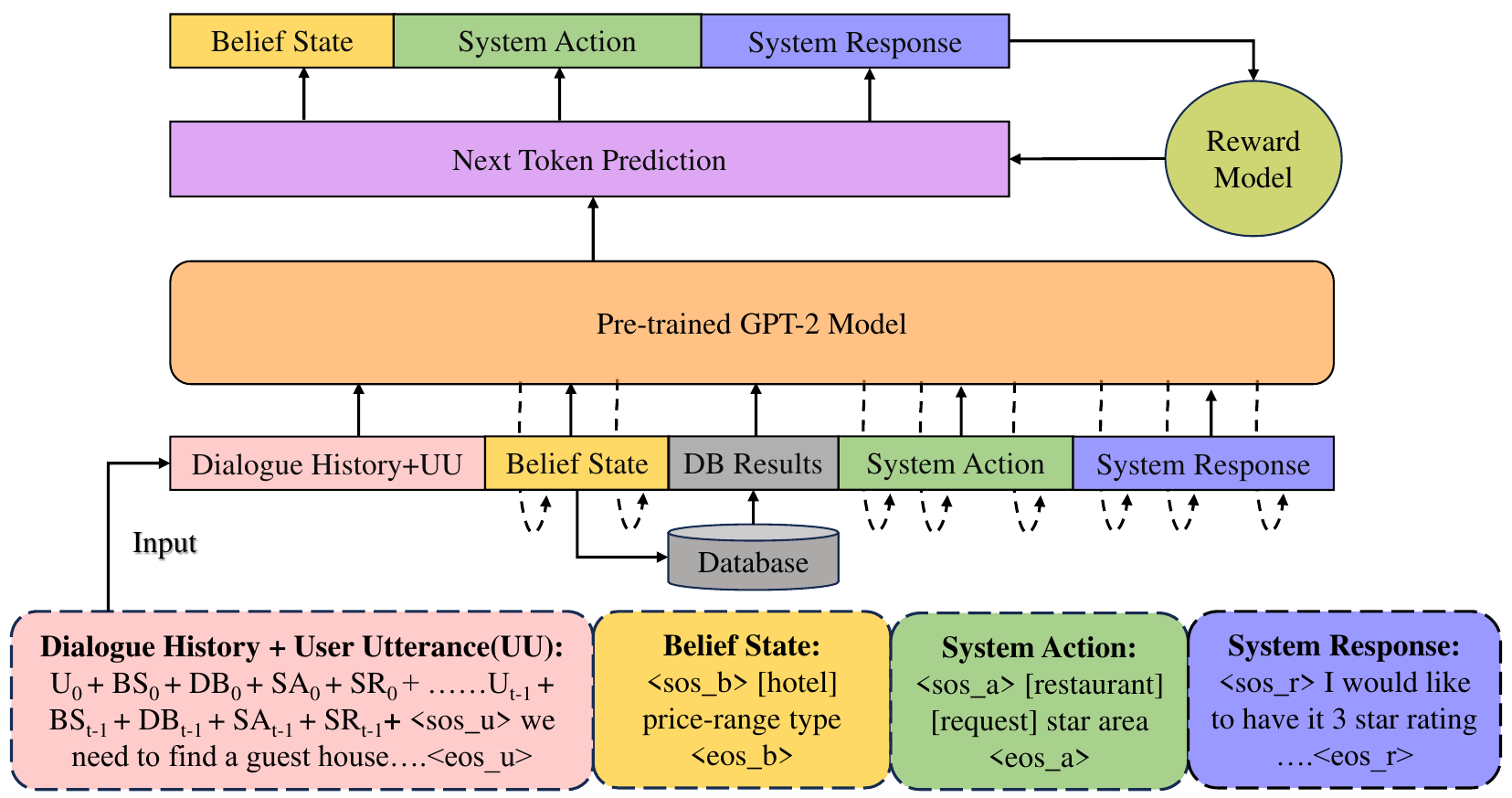} 
    \caption{The proposed architecture is fine-tuned with supervised learning which is next token prediction and reinforcement learning which is shown as a reward model.}
    \label{fig:main_diagram}
\end{figure*}

The task-oriented dialogue system is required to generate fluent and coherent natural language responses for a given user utterance. Our TOD system utilizes the pre-trained language model GPT-2 to generate the desired response. For this, our system is trained on dialogue session level which is represented as follows: ${[U_0, B_0, DB_0, A_0, R_0, U_1......U_t, B_t, DB_t, A_t, R_t]}$. A dialogue session comprises multiple components like user utterance, belief state, database search result, system act, and system response for each dialogue turn. Each component of the dialogue session is surrounded by special tokens as shown in Figure~\ref{fig:dialogue_session}. Figure~\ref{fig:main_diagram} provides a comprehensive overview of the methodology.

\subsection{Supervised fine-tuning}
Our system uses a large pre-trained unidirectional language model GPT-2 as a base model. It generates the tokens in an auto-regressive manner. The model is pre-trained on a large corpus and can generate coherent and fluent natural language. We fine-tune this model on the dialogue session level on the standard left-to-right language modeling (LM) objective. The language modeling objective maximizes the probability of the next token from the previously generated token sequence as a context. The cross-entropy loss for the language modeling objective is expressed as follows:
\begin{equation}
\label{eqn:lm}	
L_{LM} = - \sum_{i=0}^N log P(\omega_i|\omega_{0},\omega_{1},\cdot \dotsm \cdot \omega_{i-1}),
\end{equation}
where, $L_{LM}$ is an objective function, $\omega_i$ represents the next token, and $\omega_{0},\omega_{1},\cdot \dotsm \cdot \omega_{i-1}$  are the previous tokens from which the probability of next token is predicted. \textbf{$N$} represents the number of tokens in the context. 

An end-to-end task-oriented dialogue system is achieved by accomplishing sub-tasks such as belief state tracking, system act prediction, and system response generation. Each of these sub-tasks is completed through next-token prediction.

The belief states tracking sub-task records and updates the values of pre-defined slots (Ex. internet, parking, price range, etc) throughout the conversation. It ensures that the dialogue system accurately reflects the current state of the conversation and user requirements. The belief state of dialogue is generated by conditioning on the dialogue context $C_t= [U_0, B_0, DB_0, A_0, R_0, U_1 \dots, U_t ]$, which is mathematically represented as follows: 
\begin{equation}
\label{eqn:bs}	
Belief~State (B_t) = P(B_t| C_t).
\end{equation}

The belief state is generated in the following pattern: {[$domain_1$] $slot_{11}$ $value_{12}$ $slot_{12}$ $value_{12}$ [$domain_2$] $slot_{21}$ $value_{21}$}. In the generated sequence, a domain of the dialogue turn is followed by the slot and its value. This generated belief state is used to query the database and the result $DB_t$ is appended to the input sequence. Now the dialogue context becomes a sequence of $C_t$, $B_t$, and $DB_t$. Conditioned on this updated dialogue context, the TOD system generates the system action. which is mathematically represented as follows:

\begin{equation}
\label{eqn:sa}	
System~Action (A_t) = P(A_t| C_t, B_t, DB_t).
\end{equation}

The dialogue system generates the system action in the following pattern: {[$domain_1$] [$actionType_{1}$] $slot_{11}$ $slot_{12}$ \dots [$actionType_{2}$] $slot_{21}$ $slot_{22}$ \dots }. It can be seen from the generated sequence that multiple action and slot names follow the domain. This pattern is consistently produced whenever the system processes a dialogue turn. Based on all prior information $C_t$, $B_t$, $DB_t$, $A_t$, as a single sequence, the dialogue system generates the delexicalized response, which is mathematically represented as follows:
\begin{equation}
\label{eqn:sr}	
System~Response (R_t) = P(R_t| C_t, B_t, DB_t, A_t).
\end{equation}
\subsection{Reinforcement Learning}
Our goal is to steer the pre-trained language model to generate a sequence that results in high rewards. Even though the reward is scalar and non-differentiable, the reinforcement learning method updates the parameters of the pre-trained model to generate the high-rewarding token sequence. 
We use the success rate and BLEU score to calculate the reward to guide the pre-trained language model to complete the task in an end-to-end manner.
\begin{align}
\label{eqn:reward}	
{Reward}(y, \hat{y}) = & \ \alpha \times {Success}(y, \hat{y}) \notag \\
& + (1 - \alpha) \times {BLEU}(y_u, \hat{y}_u) + 1.
\end{align}
The loss function for the reinforcement learning method is calculated as the negative log probability of the generated token sequence with the highest probability multiplied by the reward. This can be expressed as follows:
\begin{equation}
\label{eqn:reinforce_loss}	
L_{RL} = - \log P(\hat{y}) \times {Reward}(y, \hat{y}).
\end{equation}
The REINFORCE loss is introduced to steer the pre-trained language model to generate the desired token sequence. Thus, the total objective of the model becomes a linear combination of $L_{LM}$ and $L_{RL}$ as follows:
\begin{equation}
\label{eqn:final_loss}	
L_{Total} = L_{LM} + \beta L_{RL}.
\end{equation}

We hyper-tune the weight, $\beta$, assigned to the loss function of the reinforcement learning. The overall process of our approach is detailed in Algorithm~\ref{alg:dctod}.

\begin{algorithm}
\caption{Offline RL for TOD System}
\label{alg:dctod}

\textbf{Input:} Dataset $D = \{(c_t, y_g )_i\}_{i=1}^{|D|}$; $y_g$ is the label for generative tasks; $e_{\text{max}}$ represent  maximum number of epochs 
\textbf{Output:} Trained Model $M$

\begin{algorithmic}[1]

\FOR{epoch $e = 1, \dots, e_{\text{max}}$}
\STATE Shuffle data $D$
\FOR{each batch $B$ in $D$}
\STATE Invoke Model $M$, using one batch of training data $B = \{(c_t, y_g)_k\}_{k=1}^{|B|}$:
\STATE \textbf{(a)} Compute the cross-entropy loss for language modeling objective 
\( LH(c_t, y_g) \) (Equation~\ref{eqn:lm}).

\STATE \textbf{(b)} Compute the reward ${Reward}(y, \hat{y})$ (Equation~\ref{eqn:reward}) and REINFORCE loss $L_{RL}$ (Equation~\ref{eqn:reinforce_loss}).

\STATE \textbf{(c)} Jointly optimize the model \( M \) by minimizing the weighted sum of loss for the language modeling task and the RL loss (Equation~\ref{eqn:final_loss}). 

\ENDFOR
% \STATE Update model parameters $\Theta$ using the joint loss.
\ENDFOR 
\end{algorithmic}
\end{algorithm}

\subsection{Decoding strategy}
Many decoding strategies are used to sample a token from the output distribution. Some are greedy decoding, beam-search decoding, top-p sampling, and contrastive search decoding methods. We use a simple greedy decoding strategy to reduce the calculation overhead at the decoding part. In the greedy decoding strategy, the token with the highest probability from the vocabulary ($V$) is selected as the next token for generation.
\begin{equation}
\hat{y}_t = \arg \max_{y_t \in V} P(y_t \mid y_1, y_2, \ldots, y_{t-1}),
\end{equation}
where: 
\begin{itemize}
    \item $\hat{y}_t$ is the generated token at step $t$
    \item $P(y_t \mid y_1, y_2, \ldots, y_{t-1})$ is the conditional probability of the token $y_t$  over the vocabulary size ($V$).
\end{itemize}

\begin{table*}[ht]
\caption{Performance of our approach in the end-to-end setting on MultiWOZ2.1 test dataset. The best results are shown in bold. *~indicates the reconstructed results.}
\begin{center}
\begin{tabular}{|c|c|c|c|c|c|}
\hline
\multirow{2}{*}{{Model}}&\multirow{2}{*}{{Pre-trained Model}}&\multicolumn{4}{|c|}{{MultiWOZ2.1}} \\
\cline{3-6} 
&\textbf{} & {{Inform Rate}}& {{Success Rate}}& {{BLEU}} & {{Combined Score}}\\
\hline
LABES~\cite{zhang-etal-2020-probabilistic} & -                   &74.50 &63.90 &16.00 &85.20 \\
SimpleTOD~\cite{hosseini2020simple} &GPT2     &84.40 &70.10 &15.01          &92.26 \\
DoTS~\cite{jeon2021domain}   &BERT-base &86.65 &74.18 &15.90          &96.31 \\
MANTOD+~\cite{zhao2023multi}  & -        &84.00 &74.8  &18.80          &98.20 \\
PPTOD~\cite{su-etal-2022-multi} &T5-base    &87.09 &79.08 &19.17          &102.26     \\
MTTOD~\cite{lee2021improving} &T5-base    &91.00 &82.10 &\textbf{21.00} &\textbf{107.50}\\
UBAR*~\cite{yang2021ubar} &distil-GPT2 &93.70 &82.00 &17.64          &105.49 \\ 
\hline

\textbf{Ours (RL)} &distil-GPT2  & \textbf{95.20} & \textbf{84.60} & 16.80 & 106.70 \\ \hline

\end{tabular}
\label{tab:main_result}
\end{center}
\end{table*}
\begin{table}[htbp]
\caption{Data statistics of train, dev, and test set based on single-domain and multi-domain dialogues of MultiWOZ2.1 dataset.}
\begin{center}
\begin{tabular}{|c|c|c|c|}
\hline
Domain &Train &Dev &Test \\ \hline
Single-Domain   &2979  &204  &226 \\
Multi-Domain  &5459  &796  &774  \\ \hline
Total     &8438  &1000  &1000  \\
\hline
\end{tabular}
\label{tab:dataset}
\end{center}
\end{table}
\section{Experimental Settings}
\subsection{Baselines:} We compare our method, with several strong baselines:
\textbf{LABES-S2S}~\cite{zhang-etal-2020-probabilistic}: The method proposes a probabilistic dialogue approach where belief states are represented as discrete latent variables and jointly modeled with system responses based on user inputs.

\textbf{SimpleTOD}~\cite{hosseini2020simple}: This approach uses a pre-trained GPT2 language model for TOD system in end-to-end setting. They fine-tune this model on the dialogue turn level, using one dialogue turn at a time within the dialogue context. The dialogue context contains the oracle belief state instead of the generated one.  

\textbf{DoTS}~\cite{jeon2021domain}: This approach works to reduce the dialogue context length which can go longer with each dialogue turn. The model tracks the domain along with the belief state and adds this into the dialogue context to limit the length of the input sequence.

\textbf{PPTOD}~\cite{dathathriplug}: This method uses extra corpora similar to MultiWOZ dataset during the pre-training stage to enhance the model performance. They use the T5-base as a base model for the end-to-end TOD system.

\textbf{MTTOD}~\cite{lee2021improving}: MTTOD system is modeled in an end-to-end setting. They use the span prediction as an auxiliary task to improve the internal presentation of the pre-trained model and use the T5 pre-trained model as a base model. 

\textbf{UBAR}~\cite{yang2021ubar}: This approach uses a pre-trained GPT2 language model as a base model which is fine-tuned on the dialogue-session level. In the dialogue-session level, dialogue context comprises, user-utterance, generated belief-state, generated dialogue-act, and dialogue response of all dialogue turns. The dialogue session is used as an input sequence. In this approach, the TOD system fully operates in an end-to-end setting. We reconstructed the results of UBAR by adapting their code.

\subsection{Dataset}
We use the MultiWOZ2.1~\cite{eric-etal-2020-multiwoz} dataset to fine-tune the pre-trained model. It is a multi-domain task-oriented dialogue dataset that contains dialogues from seven distinct domains. The domains of this dataset are as follows: hotel, restaurant, taxi, train, attraction, police, hospital. The training set of this dataset includes 8,438 dialogues spanning over seven domains, while the validation and test set each contains 1,000 dialogues, excluding the hospital and police domains. A dialogue turn of a dialogue session may also belong to more than one domain. Table~\ref{tab:dataset} shows the single-domain and multi-domain dialogues available in the dataset. 

% \begin{table}[htbp]

\subsection{Training Details}
We use a distilled version of the pre-trained GPT-2 language model as a base mode and the GPT2Tokenizer for tokenizing the sentences. The distilled version of GPT2 consists of six transformer's decoder blocks. We use the batch size of 2 with the gradient accumulation step of 16 for 50 epochs on the MultiWOZ2.1 training dataset to fine-tune the pre-trained model. The pre-trained model can take the input sequence with the length of 1024 tokens and any input sequence longer than this is truncated before being given as input to the model. We use the AdamW optimizer with a learning rate of 0.0001 for optimization and select the best-performing model on the validation set based on the evaluation metric with the highest success rate. The hyper-parameter $\alpha$ controls the trade-off between success rate and BLEU score and the hyper-parameter $\beta$ is a weight given to the RL loss function. We set $\alpha$ to 3 and $\beta$ to 0.8 after conducting a hyper-parameter search manually.

\subsection{Evaluation Metrics}
We follow the automatic evaluation metrics: inform rate, success rate, and bleu to evaluate the model in an end-to-end setting for task completion. \textbf{Inform rate} evaluates whether the system has correctly provided all entities. and \textbf{success rate} evaluates whether the system has provided all the requested information while \textbf{BLEU}~\cite{papineni2002bleu} evaluates the model on fluency and coherency of dialogue response~\cite{budzianowski2018large}. A combined score (Comb): (Inform rate + Success rate) $\times$ 0.5 + BLEU measures the overall quality of the system response~\cite{mehri-etal-2019-structured}.

\begin{table*}[htbp]
\caption{Performance comparison of our model against other baseline models when trained on full dataset and validated and tested on single-domain and multi-domain dialogues in an end-to-end setting.}
\begin{center}

\begin{tabular}{|c|c|c|c|c|c|c|}
\hline
\multirow{2}{*}{{Train Set}} &\multirow{2}{*}{{Validation Set}}&\multirow{2}{*}{{Model}}&\multicolumn{4}{|c|}{{MultiWOZ2.1}} \\
\cline{4-7} 
 &&\textbf{} & {\textit{Inform Rate}}& {{Success Rate}}& {{BLEU}} & {{Combined Score}}\\
\hline
\multirow{3}{*}{Full-set} &\multirow{3}{*}{Single-domain} & MANTOD+   &88.00 &80.4 &\textbf{17.4} &101.60 \\
% \cline{2-6} 
&& UBAR*     &97.31 &\textbf{94.17} &15.14 &110.88\\
% \cline{2-6} 
&& \textbf{Ours}      &\textbf{99.55} &\textbf{94.17}  &14.85 &\textbf{111.72} \\
\hline
\multirow{3}{*}{Full-set} &\multirow{3}{*}{Multi-domain} &MANTOD+ &83.20 &73.50 &\textbf{19.20}&97.55 \\
% \cline{2-6} 
&& UBAR* &92.66 &78.51 &18.03 &103.61\\
% \cline{2-6}
&& \textbf{Ours} &\textbf{93.95} &\textbf{81.85} &17.10 &\textbf{105.00}\\
\hline
% \multirow{3}{*}{Single-domain} &\multirow{3}{*}{Single-domain} & MANTOD+   &74.50 &63.90 &\textbf{16.00} &85.20 \\
% % \cline{2-6} 
% && UBAR*     &97.31 &94.17 &15.14 &110.88\\
% % \cline{2-6} 
% && \textbf{Ours}      &\textbf{97.31} &\textbf{80.27}  &10.28 &\textbf{99.07} \\
% \hline
% \multirow{3}{*}{Multi-domain} &\multirow{3}{*}{Multi-domain} &MANTOD+ &81.80 &72.50 &\textbf{19.80}&96.95 \\
% % \cline{2-6} 
% && UBAR* &92.66 &78.51 &18.03 &103.61\\
% % \cline{2-6}
% && \textbf{Ours} &\textbf{92.92} &\textbf{80.95} &17.60 &\textbf{104.54}\\
% \hline

% \multicolumn{4}{l}{$^{\mathrm{a}}$Sample of a Table footnote.}
\end{tabular}
\label{tab:train_on_validate_on}
\end{center}
\end{table*}

\begin{figure*}[htbp]
    \centering
    \begin{tabular}{ccc}
        \includegraphics[width=0.312\textwidth]{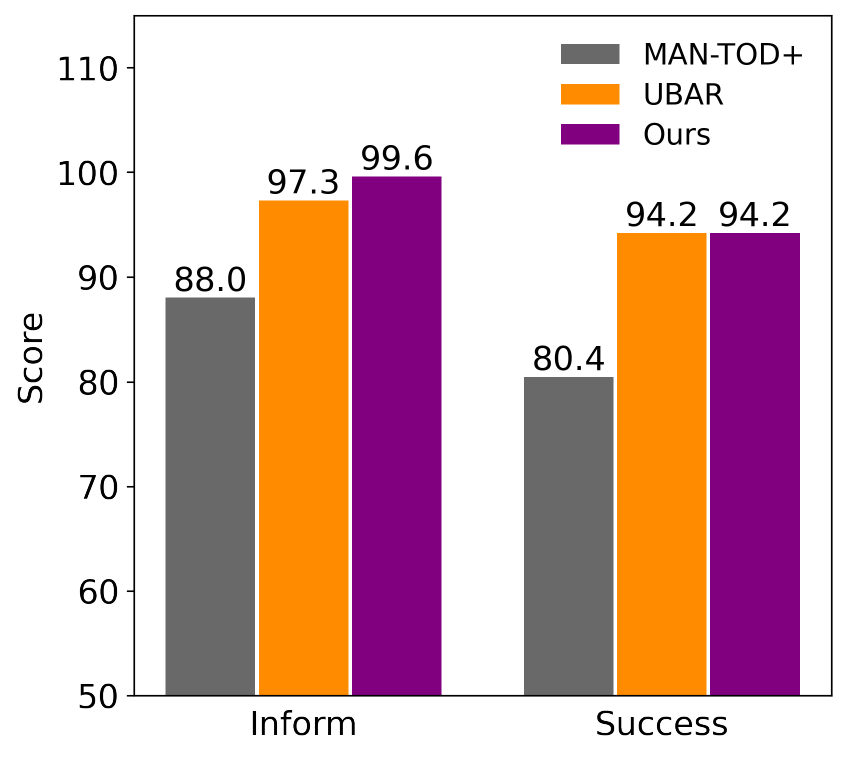} &
        \includegraphics[width=0.312\textwidth]{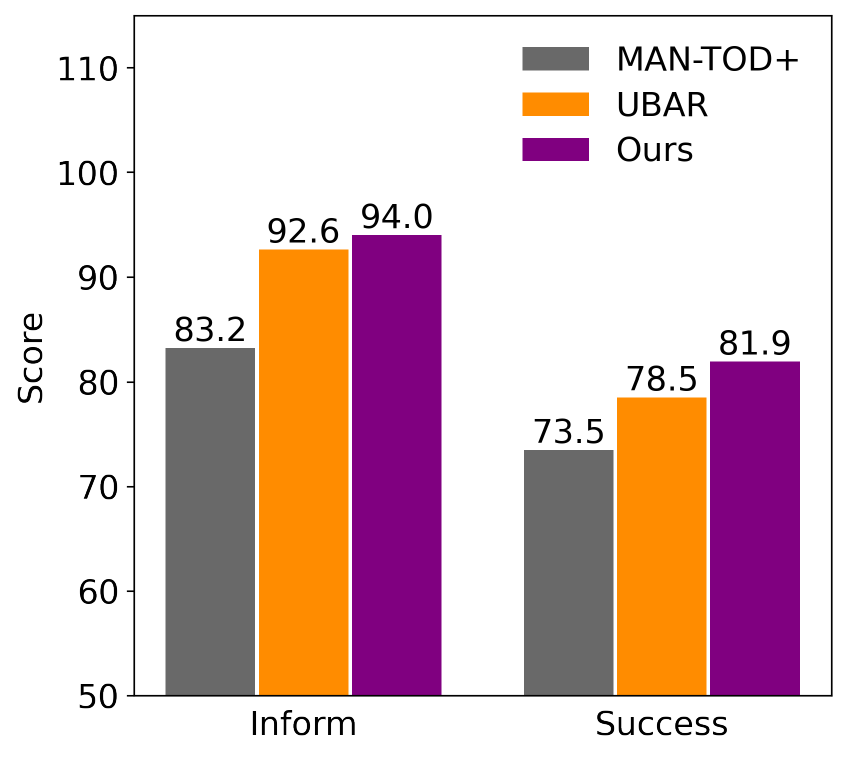} &
        \includegraphics[width=0.312\textwidth]{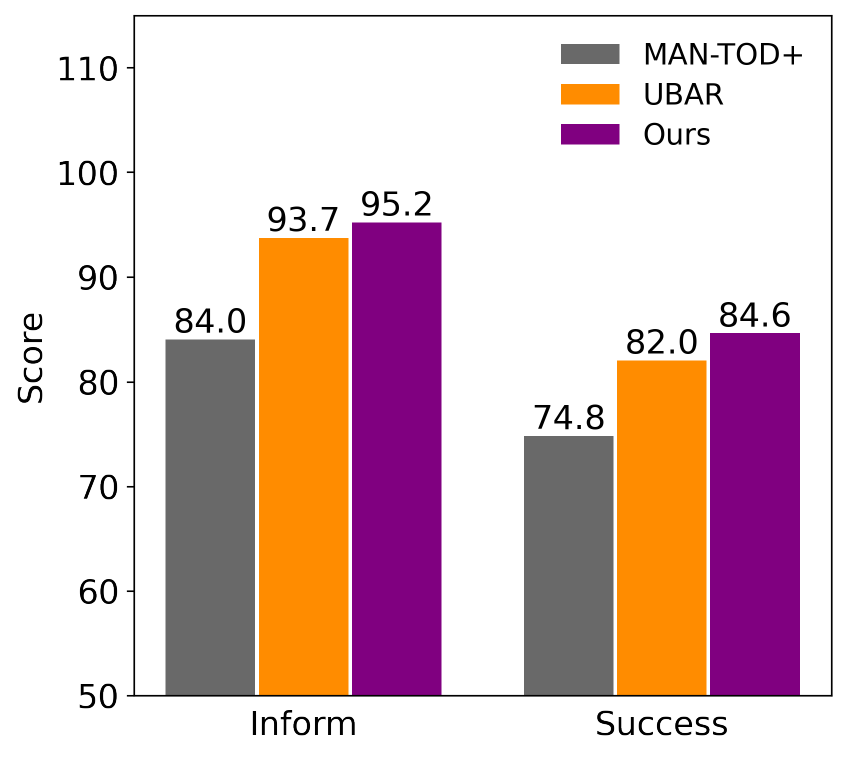} \\
        (a) Single-Domain & (b) Multi-Domain & (c) Full-Set \\
    \end{tabular}
    \caption{The proposed method and other baseline methods, are trained on the complete training dataset and evaluated on single-domain, multi-domain, and full-set. Here, we illustrate the results on the MultiWOZ2.1 test dataset. All experiments are conducted in an end-to-end setting.}
    \label{fig:three_diagrams}
\end{figure*}

\section{Results and Discussion}
\subsection{End-to-end Modeling}
Table~\ref{tab:main_result}, shows the performance of our model in an end-to-end setting on the MultiWOZ2.1 dataset. In this setting, All intermediate information of the TOD system is generated rather than relying on the oracle/ground-truth values. We use the generated belief state to query the database and the generated system acts to generate the system response.
The experimental results demonstrate that the pre-trained model, fine-tuned with a combination of offline reinforcement learning and supervised learning, significantly improves the inform rate and success rate. Results in Table~\ref{tab:main_result} show that our approach demonstrates the competitive performance with other baseline methods. Although the combined score is influenced by the BLEU, our model still surpasses the baseline on this metric. This indicates that our method performs much better on the inform rate and success rate to compensate for the comparatively low BLEU score. Specifically, it delivers improvements of 1.60\% in inform rate and 3.17\% in success rate over the baseline method, UBAR on the MultiWOZ2.1 dataset. 
\begin{figure}[ht]
    \centering
    \includegraphics[width=\columnwidth]{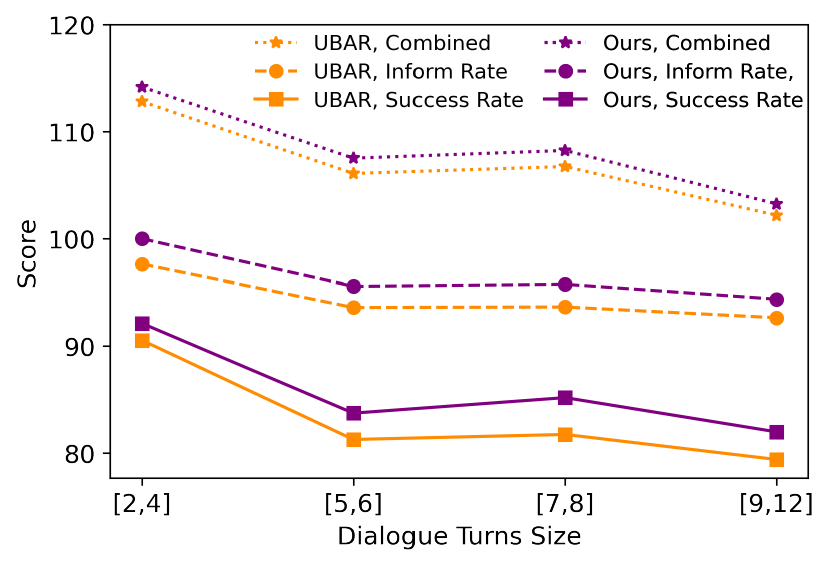}
    \caption{Performance of our system across different dialogue turns size on MultiWOZ2.1 test dataset.}
    \label{fig:dialogue_turn}
\end{figure}

% In another set of experiments, we split the training data into single-domain and multi-domain sets, and the model is trained and tested on a similar type of data. The model with the highest success rate on a specific validation set is used to report the results on the corresponding test data. We use the same training setting as we used for our best-performing model for end-to-end modeling. Table~\ref{tab:train_on_validate_on} shows the results of our experiments.

% \begin{table}[htbp]
% \caption{Performance of our model with different training and evaluation settings. \textit{Single} denotes the single-domain and \textit{Multi} denotes the multi-domain dataset}
% \begin{center}
% \begin{tabular}{|c|c|c|c|c|c|}
% \hline
% \multirow{2}{*}{{Train}}&\multirow{2}{*}{{Validation}}&\multicolumn{4}{|c|}{\textbf{MultiWOZ2.1}} \\
% \cline{3-6} 
%  &\textbf{} & \textbf{\textit{Inform}}& \textbf{\textit{Success}}& \textbf{\textit{BLEU}} & \textbf{\textit{Combined}}\\
% \hline
% {Single} & Single   &74.50 &63.90 & &85.20 \\
% \hline
% {Single} & Multi   &74.50 &63.90 & &85.20 \\
% \hline
% {Multi} & Multi   &74.50 &63.90 & &85.20 \\
% \hline
% {Multi} & Full-set   &74.50 &63.90 & &85.20 \\
% \hline
% \end{tabular}
% \label{tab:train_on_validate_on}
% \end{center}
% \end{table}

\begin{table*}[htbp] 
\caption{Performance comparison of our model with the base model(BM), UBAR, in end-to-end setting across different dialogue turns on the MultiWOZ2.1 test dataset.}
\begin{center}
\begin{tabular}{|c|c|c|c|c|c|c|c|c|}
\hline
\multirow{2}{*}{Evaluation Metrics}&\multicolumn{2}{|c|}{{[2-4]}}  & \multicolumn{2}{|c|}{{[5-6]}} & \multicolumn{2}{|c|}{{[7-8]}} & \multicolumn{2}{|c|}{{[9-12]}}  \\
\cline{2-9} 
&BM&Ours&BM&Ours&BM&Ours&BM&Ours\\
\hline 
Inform   &97.64  &100.0  &93.60  &95.57    &93.65   &95.77   &92.65  &94.38 \\
Success  &90.55  &92.13  &81.28  &83.74    &81.75   &85.19   &79.41  &81.99 \\
BLEU     &18.72  &18.11  &18.65  &17.86    &19.04   &17.76   &16.15  &15.56 \\
Combined &112.81 &114.17 &106.09 &107.52   &106.74  &108.24   &102.18 &103.25 \\
\hline
% match: 95.77  success: 85.19  bleu: 17.76    score: 108.24 (rectified)
\end{tabular}
\label{tab:turn_analysis}
\end{center}
\end{table*}

\begin{table*}[htbp]
\caption{Case Study: Comparison of responses generated by
our model with UBAR and Ground-Truth on MultiWOZ2.1 test data.}
\begin{center}
% \begin{tabular}{|c|c|}
\begin{tabular}{|p{1cm}|p{15cm}|}
\hline
\multicolumn{2}{|c|}{Dialogue Number: PMUL2933}  \\
\hline 
\textbf{User:} &no, i 'll take your recommendation. can you give me a phone number 
and postcode? oh, and how expensive is it?\\
\textbf{Ground-Truth:} &[value\_name] is [value\_price] and is absolutely gorgeous. 
the phone number is [value\_phone] and the postcode is [value\_postcode].
 would you like me to book a room? \\
\textbf{UBAR:} &i am sorry, i don't have anything that meets those criteria. would you like to try a different area? \\
\textbf{Ours:} & [value\_name] is a [value\_price] offering at postcode [value\_postcode]. 
their phone number is [value\_phone]. \\ 
\hline 
\multicolumn{2}{|c|}{Dialogue Number: PMUL4626}  \\
\hline 
\textbf{User:} &  i need to travel on tuesday to london kings cross. 		 \\
\textbf{Ground-Truth:} & [value\_id] leaves [value\_departure] for [value\_destination] on [value\_day] at [value\_leave]. would you like me to book it for you? \\ 
\textbf{UBAR:} &	[value\_id] leaves at [value\_leave] and arrives at [value\_arrive]. would you like me to book you a ticket? \\ 
\textbf{Ours:} & [value\_id] leaves [value\_departure] at [value\_leave] and arrives in [value\_destination] at [value\_arrive]. would you like to book a ticket? \\ \hline
\multicolumn{2}{|c|}{Dialogue Number: MUL1690} \\ \hline
\textbf{User:} &  can i get the price, travel time, and departure time of the 1 arriving at 18:07? 	 \\ 
\textbf{Ground-Truth:} &	the train arriving in [value\_destination] at [value\_arrive] departs at [value\_leave]. the price of the of the train ride is [value\_price]. would you like to book a train ticket? \\ 
\textbf{UBAR:} &the price is [value\_price] and the departure time is [value\_leave]. the departure time is [value\_leave]?	 \\ 
\textbf{Ours:} &[value\_id] leaves at [value\_leave] and is [value\_price] per ticket. the travel time is [value\_time]. is there anything else i can help you with?\\ \hline   

\end{tabular}
\label{tab:generated_dialogue_examples}
\end{center}
\end{table*}

\subsection{Single-Domain vs Multi-Domain}\label{AA}
To prove the effectiveness of our proposed method, we evaluate our model with two different sets of dialogues. For this, we split the test dataset into two groups, one group contains the dialogues from single-domain and another one from multi-domain. We fine-tune the pre-trained model using a complete training dataset and evaluate the model on both groups separately. Figure~\ref{fig:three_diagrams}, shows the performance comparison of our model with baseline on the evaluation metrics of inform rate and success rate. It can be seen in Table~\ref{tab:train_on_validate_on} that our model achieves comparable results on single-domain dialogues and a significant improvement on multi-domain as well as full-set of the MultiWOZ2.1 dataset with baseline. We can observe from the figure that our model achieves comparatively better results on multi-domain dialogues than on single-domain dialogues, outperforming the baseline.
\subsection{Dialogue Turn Analysis}
A task can be accomplished in any number of dialogue turns that depends on the user constraints imposed by the user during conversation. We assess the effectiveness of our model on different dialogue turn sizes on the test dataset. For this assessment, we divide the test dataset into four groups based on the dialogue turn size. The first group contains 127 dialogues with dialogue turn sizes ranging from 2 to 4. The second group includes 203 dialogues with dialogue turn sizes ranging from 5 to 6. The third group contains 378 dialogues with dialogue turn sizes between 7 to 8. 

Finally, the fourth group includes 272 dialogues with dialogue turn sizes ranging from 9 to 12. Approximately 38\% of the dialogues in the test set have dialogue turn sizes ranging from 7 to 8 this indicates that user tasks are often concluded or achieved within this number of dialogue turns. Figure~\ref{tab:turn_analysis} shows that our model achieves the highest improvement in this segment, indicating that it is highly effective when users constrain the task to 7 to 8 dialogue turns. Our model also shows superior performance on all other sets of groups. The results of the base model (BM), UBAR, and our approach are shown in Table~\ref{tab:turn_analysis}.

\begin{table}[htbp]
\caption{$\alpha$ is a trade-off between the success rate and BLEU while calculating the reward and $\beta$ is a weight to the REINFORCE loss.}
\begin{center}
\begin{tabular}{|c|c|c|c|c|c|c|c|c|}
\hline
% \multicolumn{2}{|c|}{\textbf{[2-4]}}  & \multicolumn{2}{|c|}{\textbf{[5-6]}} & \multicolumn{2}{|c|}{\textbf{[7-8]}} & \multicolumn{2}{|c|}{\textbf{[9-12]}}  \\
% \cline{2-9} 
\textbf{$\alpha$} &0.7&0.7&\textbf{0.8}&0.8&0.8&0.9&0.9\\
\hline 
\textbf{$\beta$}&3&4&\textbf{3}&4&5&3&4\\
\hline 
Inform   &93.89  &94.80  &{95.20}  &94.30    &94.90    &94.60   &95.10  \\
Success  &83.78  &83.10  &{84.60}  &82.70    &83.40    &84.00   &82.90  \\
BLEU     &17.00  &16.19  &16.80           &17.57    &17.40    &17.24   &16.05  \\
Comb &105.8 &105.1 &106.7 &106.0   &106.5  &106.5   &105.0  \\
\hline
\end{tabular}
\label{tab:hyperparameter}
\end{center}
\end{table}

\subsection{Hyper-parameter Tuning}
We perform the hyper-parameter tuning of $\alpha$ and $\beta$ manually. The hyper-parameter $\alpha$ controls the trade-off between BLEU and success rate while calculating the reward. The trade-off between success rate and BLEU is important as it helps the model to improve the success rate without losing the coherency and fluency of the generated response. The hyper-parameter $\beta$ balances the importance between the cross entropy loss and REINFORCE loss. The performance of our model on different values of $\alpha$ and $\beta$ on the MultiWOZ2.1 test dataset is given in Table~\ref{tab:hyperparameter}. 

\subsection{Case Study}
Table~\ref{tab:generated_dialogue_examples} presents examples of system responses corresponding to user utterances from the MultiWOZ2.1 test set. As a case study, we showcase the system response from a specific dialogue turn in a multi-domain conversation to demonstrate that our model delivers higher quality system responses than the baseline. In the first example, we see that UBAR fails to provide the requested information but our model provides all the requested information and closely matches the ground truth. In the second example, our approach delivers a more relevant response to the user utterance by mentioning the source and destination of the train with its identification number and leaving time from the source station. These details are absent in UBAR's response but align more closely with the ground truth response. In the third example, we can see the UBAR gives repetitive information and does not deliver all the requested information on the other side our model provides all the requested information by specifying the train identification number. Our model also gives a more affirmative and concise response than the ground truth response.

% \subsection{Ablation Study}
\section{Conclusion}
Large language models are fine-tuned using supervised learning to adapt to task-oriented dialogue systems. Such methods tend to exhibit exposure bias problems and token loss problems. In the exposure bias problem model generates texts similar to training data and in the token loss problem model generates tokens one at a time neglecting the complete context of the input. These problems deviate the model from the successful completion of the user task. In this paper, we addressed these issues by fine-tuning a pre-trained model using offline reinforcement learning along with supervised learning. We used a non-differentiable reward to steer the language model toward the successful completion of the user task and also to generate coherent and fluent system responses. We use the success rate and BLEU evaluation metrics to calculate the reward. Through experiments, we demonstrate that offline reinforcement learning effectively guides the pre-trained model towards the successful completion of the user task by enhancing the inform rate and success rate.
\bibliographystyle{ieeetr}
\bibliography{bigdata}

\begin{thebibliography}{10}

\bibitem{seneff2000dialogue}
S.~Seneff and J.~Polifroni, ``Dialogue management in the mercury flight reservation system,'' in {\em ANLP-NAACL 2000 workshop: Conversational systems}, 2000.

\bibitem{ram2018conversational}
A.~Ram, R.~Prasad, C.~Khatri, A.~Venkatesh, R.~Gabriel, Q.~Liu, J.~Nunn, B.~Hedayatnia, M.~Cheng, A.~Nagar, {\em et~al.}, ``Conversational ai: The science behind the alexa prize,'' {\em arXiv preprint arXiv:1801.03604}, 2018.

\bibitem{young2013pomdp}
S.~Young, M.~Ga{\v{s}}i{\'c}, B.~Thomson, and J.~D. Williams, ``Pomdp-based statistical spoken dialog systems: A review,'' {\em Proceedings of the IEEE}, vol.~101, no.~5, pp.~1160--1179, 2013.

\bibitem{kim2017onenet}
Y.-B. Kim, S.~Lee, and K.~Stratos, ``Onenet: Joint domain, intent, slot prediction for spoken language understanding,'' in {\em 2017 IEEE Automatic Speech Recognition and Understanding Workshop (ASRU)}, pp.~547--553, IEEE, 2017.

\bibitem{williams2014dialog}
J.~D. Williams, M.~Henderson, A.~Raux, B.~Thomson, A.~Black, and D.~Ramachandran, ``The dialog state tracking challenge series,'' {\em AI Magazine}, vol.~35, no.~4, pp.~121--124, 2014.

\bibitem{wen-etal-2015-semantically}
T.-H. Wen, M.~Ga{\v{s}}i{\'c}, N.~Mrk{\v{s}}i{\'c}, P.-H. Su, D.~Vandyke, and S.~Young, ``Semantically conditioned {LSTM}-based natural language generation for spoken dialogue systems,'' in {\em Proceedings of the 2015 Conference on Empirical Methods in Natural Language Processing} (L.~M{\`a}rquez, C.~Callison-Burch, and J.~Su, eds.), (Lisbon, Portugal), pp.~1711--1721, Association for Computational Linguistics, Sept. 2015.

\bibitem{lei2018sequicity}
W.~Lei, X.~Jin, M.-Y. Kan, Z.~Ren, X.~He, and D.~Yin, ``Sequicity: Simplifying task-oriented dialogue systems with single sequence-to-sequence architectures,'' in {\em Proceedings of the 56th Annual Meeting of the Association for Computational Linguistics (Volume 1: Long Papers)}, pp.~1437--1447, 2018.

\bibitem{liang2020moss}
W.~Liang, Y.~Tian, C.~Chen, and Z.~Yu, ``Moss: End-to-end dialog system framework with modular supervision,'' in {\em Proceedings of the AAAI Conference on Artificial Intelligence}, pp.~8327--8335, 2020.

\bibitem{zhang2020task}
Y.~Zhang, Z.~Ou, and Z.~Yu, ``Task-oriented dialog systems that consider multiple appropriate responses under the same context,'' in {\em Proceedings of the AAAI Conference on Artificial Intelligence}, pp.~9604--9611, 2020.

\bibitem{devlin-etal-2019-bert}
J.~Devlin, M.-W. Chang, K.~Lee, and K.~Toutanova, ``{BERT}: Pre-training of deep bidirectional transformers for language understanding,'' in {\em Proceedings of the 2019 Conference of the North {A}merican Chapter of the Association for Computational Linguistics: Human Language Technologies, Volume 1 (Long and Short Papers)} (J.~Burstein, C.~Doran, and T.~Solorio, eds.), (Minneapolis, Minnesota), pp.~4171--4186, Association for Computational Linguistics, June 2019.

\bibitem{raffel2020exploring}
C.~Raffel, N.~Shazeer, A.~Roberts, K.~Lee, S.~Narang, M.~Matena, Y.~Zhou, W.~Li, and P.~J. Liu, ``Exploring the limits of transfer learning with a unified text-to-text transformer,'' {\em Journal of machine learning research}, vol.~21, no.~140, pp.~1--67, 2020.

\bibitem{hosseini2020simple}
E.~Hosseini-Asl, B.~McCann, C.-S. Wu, S.~Yavuz, and R.~Socher, ``A simple language model for task-oriented dialogue,'' {\em Advances in Neural Information Processing Systems}, vol.~33, pp.~20179--20191, 2020.

\bibitem{peng-etal-2021-soloist}
B.~Peng, C.~Li, J.~Li, S.~Shayandeh, L.~Liden, and J.~Gao, ``Soloist: Building task bots at scale with transfer learning and machine teaching,'' {\em Transactions of the Association for Computational Linguistics}, vol.~9, pp.~807--824, 2021.

\bibitem{yang2021ubar}
Y.~Yang, Y.~Li, and X.~Quan, ``Ubar: Towards fully end-to-end task-oriented dialog system with gpt-2,'' in {\em Proceedings of the AAAI Conference on Artificial Intelligence}, pp.~14230--14238, 2021.

\bibitem{ham2020end}
D.~Ham, J.-G. Lee, Y.~Jang, and K.-E. Kim, ``End-to-end neural pipeline for goal-oriented dialogue systems using gpt-2,'' in {\em Proceedings of the 58th annual meeting of the association for computational linguistics}, pp.~583--592, 2020.

\bibitem{lee2021improving}
Y.~Lee, ``Improving end-to-end task-oriented dialog system with a simple auxiliary task,'' in {\em Findings of the Association for Computational Linguistics: EMNLP 2021}, pp.~1296--1303, 2021.

\bibitem{cholakov-kolev-2022-efficient}
R.~Cholakov and T.~Kolev, ``Efficient task-oriented dialogue systems with response selection as an auxiliary task,'' in {\em Proceedings of the 5th International Conference on Natural Language and Speech Processing (ICNLSP 2022)} (M.~Abbas and A.~A. Freihat, eds.), (Trento, Italy), pp.~12--18, Association for Computational Linguistics, Dec. 2022.

\bibitem{liu2018end}
B.~Liu and I.~Lane, ``End-to-end learning of task-oriented dialogs,'' in {\em Proceedings of the 2018 Conference of the North American Chapter of the Association for Computational Linguistics: Student Research Workshop}, pp.~67--73, 2018.

\bibitem{he2022galaxy}
W.~He, Y.~Dai, Y.~Zheng, Y.~Wu, Z.~Cao, D.~Liu, P.~Jiang, M.~Yang, F.~Huang, L.~Si, {\em et~al.}, ``Galaxy: A generative pre-trained model for task-oriented dialog with semi-supervised learning and explicit policy injection,'' in {\em Proceedings of the AAAI conference on artificial intelligence}, vol.~36, pp.~10749--10757, 2022.

\bibitem{liu2021pretraining}
Q.~Liu, L.~Yu, L.~Rimell, and P.~Blunsom, ``Pretraining the noisy channel model for task-oriented dialogue,'' {\em Transactions of the Association for Computational Linguistics}, vol.~9, pp.~657--674, 2021.

\bibitem{kulhanek-etal-2021-augpt}
J.~Kulh{\'a}nek, V.~Hude{\v{c}}ek, T.~Nekvinda, and O.~Du{\v{s}}ek, ``{AuGPT}: Auxiliary tasks and data augmentation for end-to-end dialogue with pre-trained language models,'' in {\em Proceedings of the 3rd Workshop on Natural Language Processing for Conversational AI} (A.~Papangelis, P.~Budzianowski, B.~Liu, E.~Nouri, A.~Rastogi, and Y.-N. Chen, eds.), (Online), pp.~198--210, Association for Computational Linguistics, Nov. 2021.

\bibitem{bang-etal-2023-task}
N.~Bang, J.~Lee, and M.-W. Koo, ``Task-optimized adapters for an end-to-end task-oriented dialogue system,'' in {\em Findings of the Association for Computational Linguistics: ACL 2023} (A.~Rogers, J.~Boyd-Graber, and N.~Okazaki, eds.), (Toronto, Canada), pp.~7355--7369, Association for Computational Linguistics, July 2023.

\bibitem{wang2022task}
W.~Wang, Z.~Zhang, J.~Guo, Y.~Dai, B.~Chen, and W.~Luo, ``Task-oriented dialogue system as natural language generation,'' in {\em Proceedings of the 45th international ACM SIGIR conference on research and development in information retrieval}, pp.~2698--2703, 2022.

\bibitem{zhang-etal-2020-probabilistic}
Y.~Zhang, Z.~Ou, M.~Hu, and J.~Feng, ``A probabilistic end-to-end task-oriented dialog model with latent belief states towards semi-supervised learning,'' in {\em Proceedings of the 2020 Conference on Empirical Methods in Natural Language Processing (EMNLP)} (B.~Webber, T.~Cohn, Y.~He, and Y.~Liu, eds.), (Online), pp.~9207--9219, Association for Computational Linguistics, Nov. 2020.

\bibitem{jeon2021domain}
H.~Jeon and G.~G. Lee, ``Domain state tracking for a simplified dialogue system,'' {\em arXiv preprint arXiv:2103.06648}, 2021.

\bibitem{zhao2023multi}
M.~Zhao, L.~Wang, Z.~Jiang, R.~Li, X.~Lu, and Z.~Hu, ``Multi-task learning with graph attention networks for multi-domain task-oriented dialogue systems,'' {\em Knowledge-Based Systems}, vol.~259, p.~110069, 2023.

\bibitem{su-etal-2022-multi}
Y.~Su, L.~Shu, E.~Mansimov, A.~Gupta, D.~Cai, Y.-A. Lai, and Y.~Zhang, ``Multi-task pre-training for plug-and-play task-oriented dialogue system,'' in {\em Proceedings of the 60th Annual Meeting of the Association for Computational Linguistics (Volume 1: Long Papers)} (S.~Muresan, P.~Nakov, and A.~Villavicencio, eds.), (Dublin, Ireland), pp.~4661--4676, Association for Computational Linguistics, May 2022.

\bibitem{dathathriplug}
S.~Dathathri, A.~Madotto, J.~Lan, J.~Hung, E.~Frank, P.~Molino, J.~Yosinski, and R.~Liu, ``Plug and play language models: A simple approach to controlled text generation,'' in {\em International Conference on Learning Representations}, 2020.

\bibitem{eric-etal-2020-multiwoz}
M.~Eric, R.~Goel, S.~Paul, A.~Sethi, S.~Agarwal, S.~Gao, A.~Kumar, A.~Goyal, P.~Ku, and D.~Hakkani-Tur, ``{M}ulti{WOZ} 2.1: A consolidated multi-domain dialogue dataset with state corrections and state tracking baselines,'' in {\em Proceedings of the Twelfth Language Resources and Evaluation Conference} (N.~Calzolari, F.~B{\'e}chet, P.~Blache, K.~Choukri, C.~Cieri, T.~Declerck, S.~Goggi, H.~Isahara, B.~Maegaard, J.~Mariani, H.~Mazo, A.~Moreno, J.~Odijk, and S.~Piperidis, eds.), (Marseille, France), pp.~422--428, European Language Resources Association, May 2020.

\bibitem{papineni2002bleu}
K.~Papineni, S.~Roukos, T.~Ward, and W.-J. Zhu, ``Bleu: a method for automatic evaluation of machine translation,'' in {\em Proceedings of the 40th annual meeting of the Association for Computational Linguistics}, pp.~311--318, 2002.

\bibitem{budzianowski2018large}
P.~Budzianowski, T.-H. Wen, B.-H. Tseng, I.~Casanueva, U.~Stefan, R.~Osman, and M.~Ga{\v{s}}i\'c, ``Multiwoz - a large-scale multi-domain wizard-of-oz dataset for task-oriented dialogue modelling,'' in {\em Proceedings of the 2018 Conference on Empirical Methods in Natural Language Processing (EMNLP)}, 2018.

\bibitem{mehri-etal-2019-structured}
S.~Mehri, T.~Srinivasan, and M.~Eskenazi, ``Structured fusion networks for dialog,'' in {\em Proceedings of the 20th Annual SIGdial Meeting on Discourse and Dialogue} (S.~Nakamura, M.~Gasic, I.~Zukerman, G.~Skantze, M.~Nakano, A.~Papangelis, S.~Ultes, and K.~Yoshino, eds.), (Stockholm, Sweden), pp.~165--177, Association for Computational Linguistics, Sept. 2019.

\end{thebibliography}

\end{document}